\documentclass{article}
\usepackage{times}
\usepackage{graphicx} 

\usepackage[utf8]{inputenc} 
\usepackage[T1]{fontenc}    
\usepackage{hyperref}       
\usepackage{url}            
\usepackage{booktabs}       
\usepackage{amsfonts}       
\usepackage{nicefrac}       
\usepackage{microtype}      
\usepackage[inline]{enumitem}
\usepackage[margin=1in]{geometry}

\usepackage{natbib}

\usepackage{algorithm}
\usepackage{algorithmic}

\usepackage{latexsym}
\usepackage{amsmath}
\usepackage{amsthm}
\usepackage{amssymb}
\usepackage{mathtools}
\usepackage{multirow}
\usepackage{bm}
\usepackage{color,soul}
\usepackage{lscape}
\usepackage{graphicx,epstopdf}
\usepackage{psfrag}
\usepackage{rotating}
\usepackage{outlines}
\usepackage[textsize=small]{todonotes}
\usepackage{subcaption}
\usepackage{textcomp}
\usepackage{booktabs}

\newcommand{\para}[1]{\vspace{2mm}\noindent\textbf{#1}:}

\newcommand{\R}{\ensuremath{\Real}}

\newcommand{\cut}[1]{}

\newcommand{\Y}{\mathbf{Y}}
\newcommand{\X}{\ensuremath{\mathbf{X}}}
\newcommand{\U}{\mathbf{U}}
\newcommand{\V}{\mathbf{V}}

\newcommand{\step}{+}
\newcommand{\sigmoid}{\sigma}

\newcommand{\msigmoid}{{m\sigma}}

\newcommand{\Round}{\text{Round}}
\newcommand{\GRR}{\text{GRR}}

\newcommand{\GRF}{\text{GRF}}
\newcommand{\Bm}{\mathcal{B}}

\newcommand{\B}{\mathbf{B}}

\newcommand{\taus}{{\boldsymbol{\tau}}}
\newcommand{\Real}{{\mathbb{R}}}

\newcommand{\link}{\ensuremath{\psi}}
\newcommand{\lparams}{\mathbf{\theta}}
\newcommand{\rank}{\mathrm{r}}
\newcommand{\apprank}{\epsilon\text{-rank}}

\newtheorem{thm:def}{Definition}[section]
\newtheorem{thm:thm}{Theorem}[section]
\newtheorem{thm:lemma}{Lemma}[section]
\newtheorem{thm:rmk}{Remark}[section]
\newtheorem{thm:corollary}{Corollary}[section]
\newtheorem{thm:pro}{Proposition}[section]

\newcommand\tab[1][1cm]{\hspace*{#1}}

\title{Compact Factorization of Matrices Using Generalized Round-Rank}

\author{
	\bf Pouya Pezeshkpour\\
	University of California\\
	Irvine, CA\\\texttt{pezeshkp@uci.edu} \\
    \and
    \bf Carlos Guestrin\\
	University of Washington \\
	Seattle, WA\\
	\texttt{guestrin@cs.uw.edu} \\
	\and
    \bf Sameer Singh \\
	University of California\\
	Irvine, CA\\
	\texttt{sameer@uci.edu} \\
}
\date{}
\begin{document}

\maketitle

\begin{abstract}
Matrix factorization is a well-studied task in machine learning for compactly representing large, noisy data. In our approach, instead of using the traditional concept of matrix rank, we define a new notion of \emph{link-rank} based on a non-linear link function used within factorization. In particular, by applying the round function on a factorization to obtain ordinal-valued matrices, we introduce \emph{generalized round-rank} ($\GRR$). We show that not only are there many full-rank matrices that are low $\GRR$, but further, that these matrices cannot be approximated well by low-rank linear factorization. We provide uniqueness conditions of this formulation and provide gradient descent-based algorithms. Finally, we present experiments on real-world datasets to demonstrate that the $\GRR$-based factorization is significantly more accurate than linear factorization, while converging faster and using lower rank representations.  
\end{abstract}

\section{Introduction}
\label{sec:intro}

Matrix factorization is a popular machine learning technique, with applications in variety of domains, such as recommendation systems~\citep{lawrence09:non-linear,salakhutdinov08:probabilistic}, natural language processing~\citep{riedel13:relation}, and computer vision~\citep{huang03:learning}.
Due to this widespread use of these models, there has been considerable theoretical analysis of the various properties of low-rank approximations of real-valued matrices, including approximation rank~\citep{alon13:the-approximate,davenport14:1-bit} and sample complexity~\citep{balcan2017optimal}.

Rather than assume real-valued data, a number of studies (particularly ones on practical applications) focus on more specific data types,
such as binary data~\citep{nickel13:logistic}, integer data~\citep{lin2009integer}, and ordinal data~\citep{koren2011ordrec,udell14:generalized}.
For such matrices, existing approaches have used different \emph{link} functions, applied in an element-wise manner to the low-rank representation~\citep{neumann16:what}, i.e. the output $\hat{Y}$ is $\link(\U^T\V)$ instead of the conventional $\U^T\V$.
These link functions have been justified from a probabilistic point of view \citep{collins01:a-generalization,salakhutdinov08:bayesian}, and have provided considerable success in  empirical settings.
However, theoretical results for linear factorization do not apply here, and thus the expressive power of the factorization models with non-linear link functions is not clear, and neither is the relation of the rank of a matrix to the link function used.

In this paper, we first define a generalized notion of rank based on the link function $\link$, as the rank of a latent matrix before the link function is applied.
We focus on a link function that applies to factorization of integer-valued matrices: the generalized round function ($\text{GRF}$), and define the corresponding generalized round-rank ($\GRR$). 
After providing background on $\GRR$, we show that there are many low-$\GRR$ matrices that are full rank\footnote{We will refer to rank of a matrix as its \emph{linear} rank, and refer to the introduced generalized rank as \emph{link}-rank.}.
Moreover, we also study the approximation limitations of linear rank, by showing, for example, that low $\GRR$ matrices often cannot be approximated by low-rank linear matrices.
We define uniqueness for $\GRR$-based matrix completion, and derive its necessary and sufficient conditions. These properties demonstrate that many full linear-rank matrices can be factorized using low-rank matrices if an appropriate link function is used.
%

We also present an empirical evaluation of factorization with different link functions for matrix reconstruction and completion.
We show that using link functions is efficient compared to linear rank, in that gradient-based optimization approach learns more accurate reconstructions using a lower rank representation and fewer training samples.
We also perform experiments on matrix completion on two recommendation datasets, and demonstrate that appropriate link function outperform linear factorization, thus can play a crucial role in accurate matrix completion.

\section{Link Functions and Generalized Matrix Rank}
\label{sec:link}

Here we introduce our notation for matrix factorization, and use it to introduce link functions and \emph{generalized link-rank}.
We will focus on the round function and round-rank, introduce their generalized versions, and present their properties.

\para{Rank Based Factorization}
Matrix factorization, broadly defined, is a decomposition of a matrix as a multiplication of two matrices. 
Accordingly, rank of a matrix $\Y\in \Real^{n\times m}$ defined as the smallest natural number $r$ such that:
$
\Y = \U \V^T, \text{or,}
\Y_{ij} = \sum_k \U_{ik}\V_{jk}
$, 
where $\U\in \Real^{n\times r}$ and $\V\in \Real^{n\times r}$. 
We use $\rank(\Y)$ to indicate the rank of a matrix $\Y$. 

\para{Link Functions and Link-Rank}
Since the matrix $\Y$ may be from a domain $\mathbb{V}^{n\times m}$ different from real matrices, link functions can be used to define an alternate factorization: 
\begin{align}
\Y = \link_\taus(\X), 
\X = \U \V^T,
\end{align}
where $\Y \in \mathbb{V}^{n\times m}$, $\link:\Real\rightarrow\mathbb{V}$ (applied element-wise), $\X\in \Real^{n\times m}$, $\U\in \Real^{n\times r}$, $\V\in \Real^{n\times r}$, and $\taus$ represent parameters of the link function, if any. 
Examples of link functions that we will study in this paper include the \emph{round} function for binary matrices, and its generalization to ordinal-valued matrices.
Link functions were introduced for matrix factorization by \citet{singh08:a-unified}, consequently \citet{udell14:generalized} presented their generalization to loss functions and regularization for \emph{abstract data types}.

\begin{thm:def}\label{thm:rank}
    Given a matrix $\Y$ and a link function $\link_\taus$ parameterized by $\taus$, the \textbf{link-rank} $\rank_\link$ of $\Y$ is defined as the minimal rank of a real-matrix $\X$ such that,  $\Y = \link_\taus(\X)$,
    \begin{equation} \label{eq:rank}
        \rank_\link(\Y) = \min_{\X\in\Real^{n\times m}, \taus} \left\{ \rank(\X); \Y = \link_\taus(\X) \right\}
    \end{equation}
\end{thm:def}
Note that with $\link\equiv I$, i.e. $\link(x)=x$, $\rank_\link(\Y)=\rank(\Y)$. 

\para{Sign and Round Rank}
If we consider the $\text{sign}$ function as the link function, where $
\text{sign}(x)= \{ 0 \text{ if } x<0, 1 \text{ o.w.} \}$
(applied element-wise to the entries of the matrix), the link-rank defined above corresponds to the well-known $\text{sign-rank}$ for binary matrices~\citep{neumann2015some}: 
\[
\text{sign-rank}(\Y) = \min_{\X\in\Real^{n\times m}} \left\{ \rank(\X); \Y = \text{sign}(\X) \right\}.
\]
A variation of the $\text{sign}$ function that uses a threshold $\tau$, $\Round_\tau(x)=\{ 0 \text{ if } x<\tau, 1 \text{ o.w.} \}$ when used as a link function results in the $\text{round-rank}$ for binary matrices, i.e.
\[
\text{round-rank}_{\tau}(\Y) = \min_{\X\in\Real^{n\times m}} \left\{ \rank(\X); \Y = \Round_{\tau}(\X) \right\},
\] %
as shown in ~\citet{neumann2015some}. 
Thus, our notion of \emph{link-rank} not only unifies existing definitions of rank, but can be used for novel ones, as we will do next.

\para{Generalized Round-Rank ($\GRR$)}
\label{sec:rrabk}
Many matrix factorization applications use ordinal values, i.e $\mathbb{V}=\{0,1,\ldots,N\}$. 
For these, we define generalized round function ($\text{GRF}$): 
\begin{align}
\text{GRF}_{\tau_1,...,\tau_N}(x)=
\begin{cases}
               0\tab\tab x \leq \tau_1\\
               1\tab\tab \tau_1 < x \leq \tau_2\\
               \vdots\\
               N-1\tab \tau_{N-1} < x \leq \tau_{N}\\
               N\tab\tab \text{o.w.}
            \end{cases}
\end{align}
where its parameters $\taus\equiv\{\tau_1,...,\tau_N\}$ are thresholds (sorted in ascending order). 
Accordingly, we define \emph{generalized round-rank} ($\GRR$) for any ordinal matrix $\Y$ as: 
\[
\GRR_{\taus}(\Y)=\min_{\X\in\Real^{n\times m}} \left\{ \rank(\X); \Y = \text{GRF}_{\taus}(\X) \right\}
.
\]
Here, we are primarily interested in exploring the utility of $\GRR$ and, in particular, compare the representation capabilities of low-$\GRR$ matrices to low-linear rank matrices.
To this end, we present the following interesting property of $\GRR$. 

\begin{thm:thm}
\label{thm:thresh}
For a given matrix $\Y \in \{0,\ldots,N\}^{n \times m}$, let's assume $\taus^*$ is the set of optimal thresholds, i.e. $ \GRR_{\taus^{\star}}(Y)=\text{argmin}_{\taus}\GRR_{\taus}(Y)$, then for any other $\taus'$: 
\begin{align}
\label{thresh}
\GRR_{\taus'}(\Y) \leq N\times\GRR_{\taus^{\star}}(\Y)+1
\end{align}
\begin{proof}
We provide a sketch of proof here, and include the details in the appendix.
%
We can show that the GRR can change at most by $1$ if we add a constant to all the thresholds and does not change at all if all the thresholds are multiplied by a constant. 
Further, we show that there exist $\epsilon_i$ for every $i \in \{1,...,N-1\}$ such that shifting $\tau_i$ by $\epsilon_i$ does not change the GRR. 
These properties provide a bound to the change in GRR between any two sets of thresholds. 
\end{proof}
\end{thm:thm}
This theorem shows that even though using a fixed set of thresholds is not optimal, the rank is still bounded in terms of $N$, and does not depend on the size of the matrix ($n$ or $m$). Other complementary lemmas are provided in appendix.
\begin{thm:rmk}
The upper bound in the theorem \ref{thm:thresh} matches the upper bound found in \citet{neumann16:what} for the case where $N=1$,
$
\GRR_{\tau '}(\Y) \leq  \GRR_{\tau^*}(\Y)+1
$. 
\end{thm:rmk}

\section{Comparing Generalized Round Rank to Linear Rank}
\label{sec:GRR vs LR}

Matrix factorization (MF) based on linear rank has been widely used in lots of machine learning problems like matrix completion, matrix recovery and recommendation systems. The primary advantage of matrix factorization is its ability to model data in a compact form. 
Being able to represent the same data accurately in an even more compact form, specially when we are dealing with high rank matrices, is thus quite important.
Here, we study specific aspects of exact and approximate matrix reconstruction with $\GRR$. 
In particular, we introduce matrices with high linear rank but low $\GRR$, and demonstrate the inability of linear factorization in approximating many low-$\GRR$ matrices.

\subsection{Exact Low-Rank Reconstruction}
To compare linear and $\GRR$ matrix factorization, here we identify families of matrices that have high (or full) linear rank but low (or constant) $\GRR$.
Such matrices demonstrate the primary benefit of GRR over linear rank: factorizing matrices using GRR can be significantly beneficial. 

As provided in \citet{neumann2015some} for round-rank (a special case of $\GRR$), $\GRR_{\taus}(\Y)\leq r(\Y)$ for any matrix $\Y\in\mathbb{V}^{n\times m}$. 
More importantly, there are many structures that lower bound the linear rank of a matrix.
For example, if we define the upper triangle number $n_U$ for matrix $\Y\in \mathbb{V}^{n\times n}$ as the size of the biggest square block which is in the form of an upper triangle matrix, then $
\rank(\Y)\geq n_U
$. 
If we define the identity number $n_I$ similarly, then
$
\rank(\Y)\geq n_I
$, 
and similarly for matrices with a band diagonal submatrix. 
None of these lower bounds that are based on identity, upper-triangle, and band-diagonal structures apply to $\GRR$. 
In particular, as shown in \citet{neumann2015some}, identity matrices (of any size) have a constant round-rank of $2$, upper triangle matrices have round-rank of $1$, and band diagonal matrices have round-rank of $2$ (which also holds for GRR). 
Moreover, we provide another lower bound for linear rank of a matrix, which is again not applicable to $\GRR$.   
\begin{thm:thm}
	If a matrix $\Y\in\R^{n\times m}$ contains $k$ rows, $k\leq n,k\leq m$, such that $R=\{Y_{R_1},...,Y_{R_k}\}$, two columns $C=\{j_0,j_1\}$, and:
	\begin{enumerate}[nosep]
	\item rows in $R$ are distinct from each other, i.e, $\forall i,i'\in R, \exists j,Y_{ij}\neq Y_{i'j}$,
	\item columns in $C$ are distinct from each other, i.e, $\exists i,Y_{ij_0}\neq Y_{ij_1}$, and
	\item matrix spanning $R$ and $C$ are non-zero constants, w.l.o.g. $\forall i\in R,Y_{ij_0}=Y_{ij_1}=1$,
	\end{enumerate}
	then $\rank(\Y)\geq k$. (See appendix for the proof)
\end{thm:thm}

So far, we provide examples of high linear-rank structures that do not impose any constraints on $\GRR$.
We now provide the following lemma that, in conjunction with above results, indicates that lower bounds on the linear rank can be really high for matrices if they contain low-GRR structures (like identity and upper-triangle), while the lower bound on $\GRR$ is low.
\begin{thm:lemma}
For any matrix $A$, if there exists a submatrix $A'$ in a way that $\rank(A')=R$ and $\GRR_{\taus}(A')=r$, then $\GRR_{\taus}(A)\geq  r$ and $\rank(A)\geq R$. 
\begin{proof}
If we consider the linear rank as the number of independent row (column) of the matrix, consequently having a rank of $R$ for submatrix $A'$ means there exist at least $R$ independent rows in matrix $A$. Using this argument we can simply prove above inequalities.
\end{proof}
\end{thm:lemma}


\subsection{Approximate Low-Rank Reconstruction}

Apart from examples of high linear-rank matrices that have low GRR, we can further show that many of these matrices cannot even be \emph{approximated} by a linear factorization.
In other words, we show that there exist many matrices for which not only their linear rank is high, but further, that the linear rank approximations are poor as well, while their low GRR reconstruction is perfect.
In order to measure whether a matrix can be approximated well, we describe the notion of approximate rank (introduced by \citet{alon13:the-approximate}, we rephrase it here in our notation).
\begin{thm:def}
Given $\epsilon$, \textbf{approximate rank} of a matrix $\X$ is:\\
$
\apprank(\X) = \min\{\rank(\X'): \X'\in\Real^{n\times m}, ||\X-\X'||^2_F\le\epsilon\}
$
\end{thm:def}
We extend this definition to introduce the generalized form of approximate rank as follows:
\begin{thm:def}
Given $\epsilon$ and a link function $\link$ (e.g. GRF), the \textbf{generalized approximate rank} of a matrix $\Y$ is defined as:
\hspace{5mm}
$
\apprank_\link(\Y) = \min\{\rank_\link(\Y'): \Y'\!\in\!\mathbb{V}^{n\times m}, ||\Y-\Y'||^2_F\le\epsilon\}
$. 
\end{thm:def}


For an arbitrary matrix, we can evaluate how well a linear factorization can approximate it using SVD, i.e.:
\begin{thm:thm}
\label{thm:pca}
For a matrix $\X=\U\Sigma\V^T$, where $\mathrm{diag}(\Sigma)$ are the singular values, and $\U$ and $\V$ are orthogonal matrices, then $\sum_{i=k+1}^n|\Sigma_{ii}|^2 = \min_{\Y,\rank(\Y)=k}||\X-\Y||^2_F$.
\begin{proof}
This was first introduced in ~\citet{eckart1936approximation}, and recently presented again in ~\citet{udell14:generalized}.
We omit the detailed proof, but the primary intuition is that the PCA decomposition minimizes the Frobenius norm, and $\Y=\U'\V'$, with $\U'=\U\Sigma^{\frac{1}{2}}$ and $\V'=\Sigma^{\frac{1}{2}}\V^T$.
\end{proof}
\end{thm:thm}
For an arbitrary binary matrix $\Y$, recall that $\Round_{\tau=0}(\Y)$ is equal to $\text{sign-rank}(\Y)$. 
Using above theorem, we want to show that there are binary matrices that cannot be approximated by low linear-rank matrices (for non-trivial $\epsilon$), but can be approximated well by low round-rank matrices. %
Clearly, these results extend to ordinal matrices and their $\GRR$ approximations, the generalized form of binary case. 

\begin{figure}
    \centering
        \includegraphics[width=0.7\columnwidth]{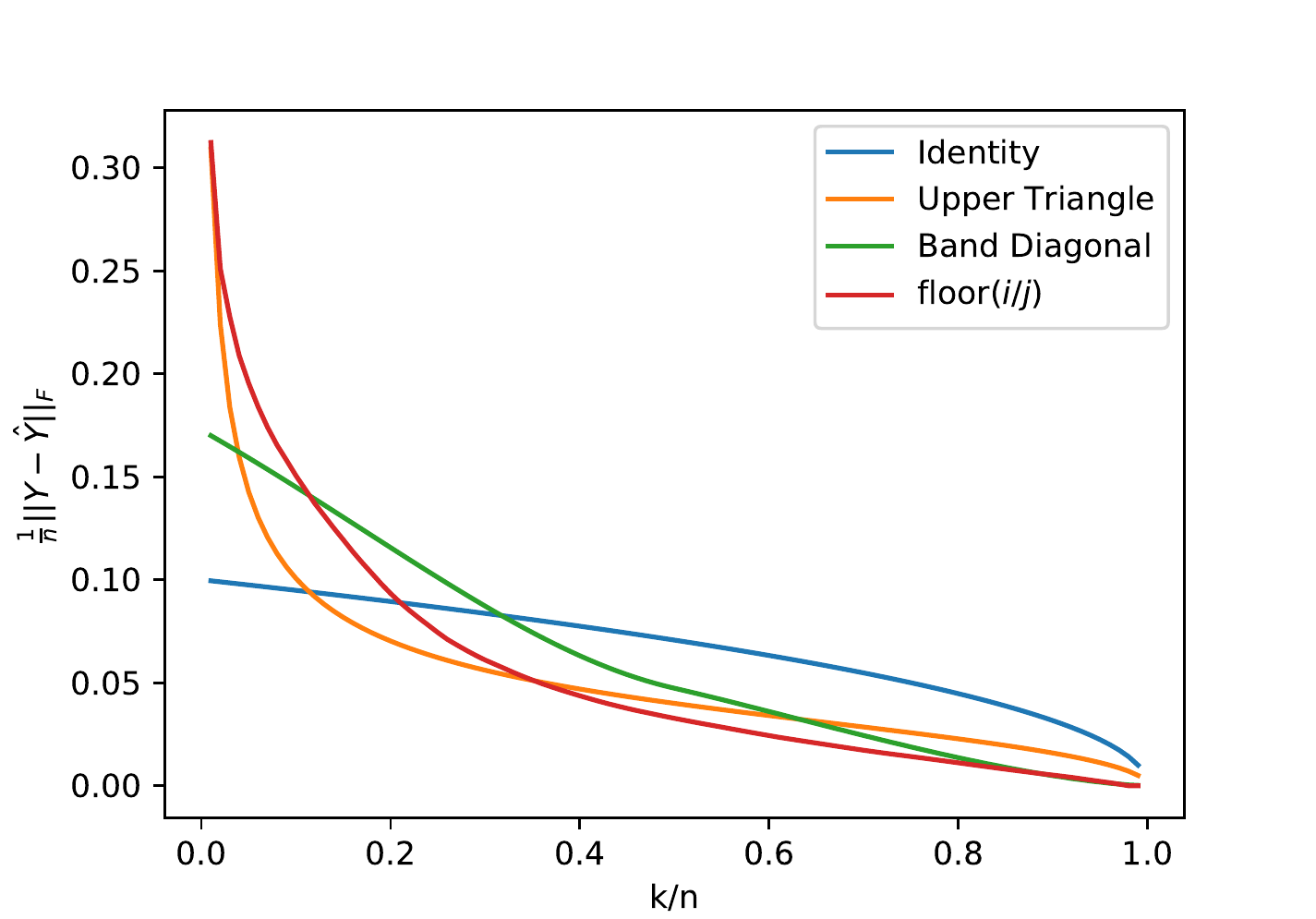}
    	\caption{Comparison of the optimal linear factorization approximation as the rank $k$ is varied for a number of matrices (of size $n\times n$), demonstrating that linear factorization is unable to approximate these matrices with low-rank. All of these matrices have a \emph{constant} generalized round-rank ($\leq 2$).}
    	\label{fig:pca_plot}
\end{figure}

Let us consider $\Y$, the identity binary matrix of size $n$, for which the singular values of $\Y$ are all $1$s. 
By using Theorem~\ref{thm:pca}, any linear factorization $\Y'$ of rank $k$ will have $||\Y-\Y'||_F^2\geq(n-k)$.
As a result, the identity matrix cannot be approximated by \emph{any} rank-$k$ linear factorization for $\epsilon<{n-k}$. 
On the other hand, such a matrix can be reconstructed exactly with
a rank $2$ factorization if using the round-link function, since $\text{round-rank}(\Y)=2$.
%
%
%
%
In Figure~\ref{fig:pca_plot}, we illustrate a number of other such matrices, i.e. they can be exactly represented by a factorization with $\GRR$ of $2$, but cannot be approximated by any compact linear factorization.
\section{Matrix Completion with Generalized Round-Rank Factorization}
\label{sec:SC}

So far, we show that there are many matrices that cannot be represented compactly using conventional matrix factorization (linear), either approximately or exactly, whereas they can be reconstructed using compact matrices when using $\GRF$ as the link function.
In this section, we study properties of \emph{completion} of ordinal-valued matrices based on $\GRF$ (and the notion of rank from $\GRR$).
In particular, given a number of noise-free observations $\Omega$ from $\Y\in\{0,\ldots,N\}^{n\times m}$ and its $\GRR(\Y)=r, r\ll\min(n,m)$, the goal here is to identify $\U\in\Real^{n\times r},\V\in\Real^{m\times r}$ such that $\GRF(\U\V^T)$ completes the unobserved entries of $\Y$ accurately.


\subsection{Theoretical Results for Uniqueness}

Uniqueness in matrix completion is defined as the minimum number of entries required to recover the matrix $\Y$ with high probability, assuming that sampling of the set of observed entries is based on an specific distribution. 
%
To obtain uniqueness in $\GRR$ based factorization, we first need to introduce the interval matrix $\mathbf{\bar{X}}$. 
Based on definition of generalized round function ($\text{GRF}$) and a set of fixed thresholds, we define matrix $\mathbf{\bar X}$ to be a matrix with interval entries calculated based on entries of matrix $\Y$ and thresholds ($\tau_1,...\tau_N$). 
As an example, if an entry $\Y_{ij}$ is $k \in \{0,...,N\}$, $\mathbf{\bar X_{ij}}$ would be equal to the interval $[\tau_k,\tau_{k+1}]$. 
When entries of $\Y$ are equal to $0$ or $N$, w.l.o.g. we assume the corresponding entries in matrix $\mathbf{\bar{X}}$ are bounded. Thus, each one of matrix $\mathbf{\bar X}$'s entries must be one of the $N+1$ possible intervals based on $\text{GRF}$'s thresholds.
\begin{thm:def}
 A target matrix $\Y \in \{0,\ldots,N\}^{n \times m}$ with 1)~observed set of entries $\Omega=\{(i,j), \Y_{ij} \text{is observed}\}$, 2)~set of known thresholds ($\tau_1,...\tau_N$), and 3)~$\GRR_{\tau_1,...,\tau_N}(\Y)=r$, is called uniquely recoverable, if we can recover its unique interval matrix $\mathbf{\bar X}$ with high probability.    
\end{thm:def}

Similar to $\mathbf{\bar X}$, we introduce $\mathcal{X}^{\star}$ to be a set of all matrices that satisfy following two conditions: 1)~For the observed entries $\Omega$ of $\Y$, $\Y_{ij}=\text{GRF}_{\tau_1,...,\tau_N}(\X^{\star}_{ij})$, and 2)~linear rank of $\mathcal{X}^{\star}$ is $r$. 
If we consider a matrix $\X \in \mathcal{X}^{\star}$ then for an arbitrary entry $\X_{ij}$ we must have $\X_{ij} \in \mathbf{\bar X}_{ij}$, where $\mathbf{\bar X}_{ij}$ is an interval containing $\X_{ij}$. 
Given a matrix $\X\in\mathcal{X}^{\star}$, the uniqueness conditions ensure that we would be able to recover $\mathbf{\bar X}$, using which we can uniquely recover matrix $\Y$. 


In the next theorems, we first find the necessary condition on the entries of matrix $\X$ for satisfying uniqueness of matrix $\Y$. Then, we derive the sufficient condition accordingly.
In our calculations, we assume the thresholds to be fixed and our target matrix $\Y$ be noiseless, and further, there is at least one observed entry in every column and row of matrix $\Y$. 

\begin{thm:thm} (Necessary Condition)
For a target matrix $\Y \in \mathbb{V}^{n \times m}$ with few observed entries and given $\GRR(\Y)=r$, we consider set of $\{\Y_{i_1j},...,\Y_{i_rj}\}$ to be the $r$ observed entries in an arbitrary column $j$ of $Y$. 
Given any matrix $\X \in \mathcal{X}^{\star}$, $\X=\U\times \V^{T}$, and taking an unobserved entry $\Y_{ij}$, we define $a_{i_kj}$ as: 
$
\U_i=\sum_{k=1}^{r} a_{i_kj} \U_{i_k} 
$, where $\U_d$ ($d \in \{1,...,n\}$) is the $d^\text{th}$ row of matrix $\U$ and $i_k$ represents the index of observed entries in $j$th column.
Then, the necessary condition of uniqueness of $\Y$ is:
\begin{align}
\sum_{k=1}^{r}\left | a_{i_kj} \right | \leq  \epsilon\left(\frac{T_\text{min}}{T_\text{max}}\right)
\end{align}
Where $r=\GRR(\Y)$, $T_\text{min}$ and $T_\text{max}$ are the length of smallest and largest intervals and $\epsilon$ is a small constant. 
\begin{proof}
We only provide a sketch of proof here, and include the details in the appendix. To achieve uniqueness we need to find a condition in which for any column of $\X$, by changing respective row of $\V$, while the value of observed entries stay in the respected intervals, the value of unobserved ones wouldn't change dramatically which result in moving to other intervals. To do so, we will calculate the maximum of the possible change for an arbitrary unobserved entry of column $j$ in matrix $\Y$. To calculate this maximum for any unobserved entry $\Y_{ij}$, we consider the row $\U_i$ as a linear combination of linearly independent rows of $\U$ (which are in respect to observed entries of $\Y$ in column $j$). Then, by finding the maximum possible change for observed entries in column $j$, based on their respective intervals, we find mentioned boundary for achieving the uniqueness.
 \end{proof}
\end{thm:thm}
The same condition is necessary for matrix $\V$ as well. The necessary condition must be satisfied for all columns of matrix $\X$. 
Moreover, if the necessary condition is not satisfied, we cannot find a unique matrix $\X$, and hence a unique completion, i.e. $\Y=\text{GRF}_{\tau_1,...,\tau_N}(\X)$ where $\X \in \mathcal{X}^{\star}$.


\begin{thm:thm} (Sufficient Condition)
Using above necessary condition, for any unobserved entry $\Y_{ij}$ of matrix $\Y$ we define $\bar\epsilon$ as minimum distance of $\X_{ij}$ with its respected interval's boundaries. Then, we will have the following inequality as sufficient condition of uniqueness:
\begin{align}
\bar\epsilon \geq \max \left(\sum_{k=1}^{r}a_{i_kj}\epsilon_{i_k}^{\text{sign}(a_{i_kj})} , \left | \sum_{k=1}^{r}a_{i_kj}\epsilon_{i_k}^{-\text{sign}(a_{i_kj})}  \right |\right)
\end{align}
where $r$ and $a_{i_kj}$ are defined as before, $\epsilon_{i_kj}^{+}$ is defined as the distance of $\X_{i_kj}$ to its upper bound, and $\epsilon_{i_kj}^{-}$ is defined as negative of the distance of $\X_{i_kj}$ to its lower bound.
\end{thm:thm}
Above sufficient condition is a direct result of necessary condition proof. 
Although not tight, it guarantees the existence of unique $\mathbf{\bar X}$, and thus the complete matrix $\Y$.

\subsection{Gradient-Based Algorithm for $\GRR$ Factorization}

Although previous studies have used many different paradigms for matrix factorization, such as alternating minimization \citep{hardt2014understanding,jain2013low} and adaptive sampling~\citep{krishnamurthy2013low}, 
stochastic gradient descent-based (SGD) approaches have gained widespread adoption, in part due to their flexibility, scalability, and theoretical properties~\citep{de2014global}.
For linear matrix factorization, a loss function that minimizes the squared error is used, i.e. $L_{\text{linear}}=\sum (Y_{ij}- U_iV_j)^2$, where the summation is over the observed entries.
In order to prevent over-fitting, $L_2$ regularization is often incorporated.



\para{Round}
We extend this framework to support $\GRR$-based factorization by defining an alternate loss function. 
In particular, with each observed entry $Y_{ij}$ and the current estimate of $\taus$, we compute the $b_{ij}^\downarrow$ and $b_{ij}^\uparrow$ as the lower and upper bounds for $X_{ij}$ with respect to the $\GRF$.
Given these, we use the following loss, $L_{\text{Round}}=\sum (b_{ij}^\downarrow - U_iV_j)_++( U_iV_j - b_{ij}^\uparrow)_+$, where $(.)_+=\max(.,0)$. Considering the regularization term as well, we apply stochastic gradient descent as before, computing gradients using a differentiable form of $\max$ with respect to $\U$, $\V$, and $\taus$.

\para{Multi-Sigmoid}
Although the above loss captures the goal of the $\GRR$-based factorization accurately, it contains both discontinuities and flat regions, and thus is difficult to optimize.
Instead, we also propose to use a smoother and noise tolerant approximation of the $\GRF$ function.
The sigmoid function, $\sigma(x)=\frac{1}{1+e^{-x}}$, for example, is often used to approximate the $\text{sign}$ function.
When used as a link function in factorization, we can further show that it approximates the $\text{sign-rank}$ well. 
\begin{thm:thm}
For any $\epsilon>0$ and matrix $\Y$, $\text{sign-rank}(\Y) = \apprank_\sigmoid(\Y)$. (See appendix for the proof)

\end{thm:thm}
We can similarly approximate $\GRF$ using a sum of sigmoid functions that we call $\text{Multi-sigmoid}$ defined as $\link^\msigmoid_{\taus}(x)=\sum_{d=1}^{N}\sigma(x-\tau_d)$, for which the above properties also hold.
The resulting loss function that minimizes the squared error is $L_{\text{multi-sigmoid}}=\sum(Y_{ij}-\link^\msigmoid_\taus(U_iV_j))^2$.

In our experiments, we evaluate both of our proposed loss functions, and compare their relative performance.
%
We study variations in which the thresholds $\taus$ are either pre-fixed or updated (using $\frac{\partial}{\partial\taus}L$) during training.
All the parameters of the optimization, such as learning rate and early stopping, and the hyper-parameters of our approaches, such as regularization, are tuned on validation data.

\section{Experiments}
\label{sec:results}

In this section we evaluate the capabilities of our proposed $\GRR$ factorization relative to linear factorization first through variety of simulations, followed by considering \emph{smallnetflix} and \emph{MovieLens 100K}\footnote{The codes available at: \url{https://github.com/pouyapez/GRR-Matrix-Factorization}} datasets. 
Unless otherwise noted, all of evaluations are based on Root Mean Square Error (RMSE).

\begin{figure*}[tb]
	\centering
	\begin{subfigure}{0.31\textwidth}
	  \centering
	  \includegraphics[width=\textwidth]{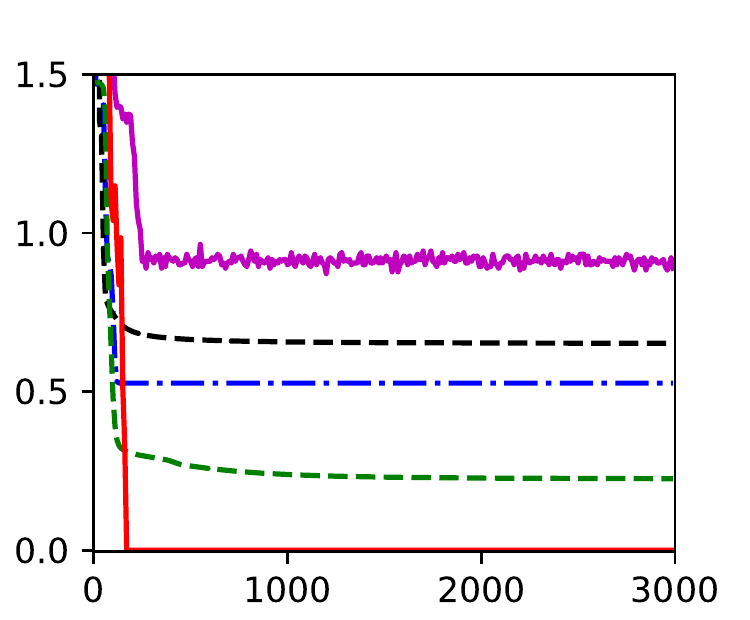}
	  \caption{Random matrix, k=2}
		\end{subfigure}
		\quad
		\begin{subfigure}{0.31\textwidth}
	  	\centering
	  	\includegraphics[width=\textwidth]{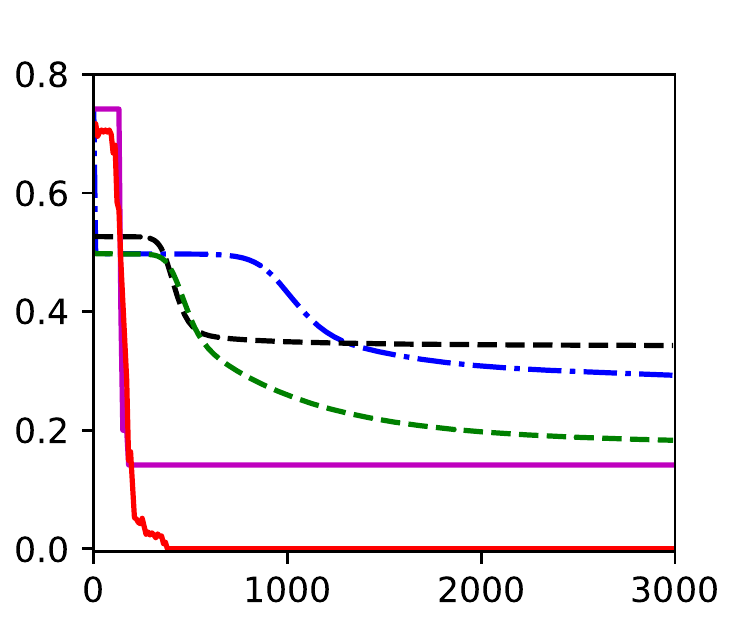}
	  	\caption{Upper Triangle matrix, k=1}
		\end{subfigure}
		\quad
		\begin{subfigure}{0.31\textwidth}
	  \centering
	  \includegraphics[width=\textwidth]{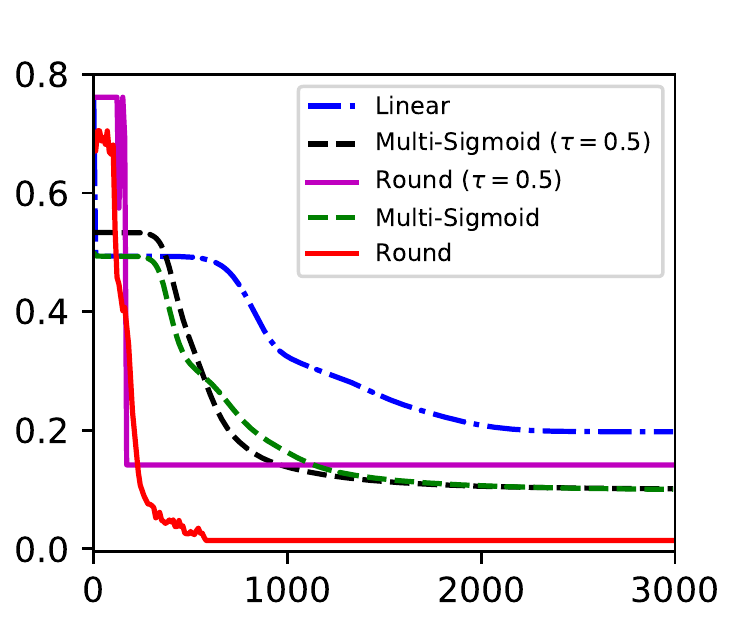}
	  \caption{Band Diagonal matrix, k=2}
	\end{subfigure}
	\caption{\textbf{Matrix Recovery:} Synthetic matrices that are reconstructed using their $k$-dimensional factorization with different representations. We plot RMSE of the reconstruction vs the number of training iterations, demonstrating the efficiency of $\GRR$-based methods, especially without fixed thresholds.}
	\label{fig:fullrec}
\end{figure*}

\paragraph{Matrix Recovery}
We first consider the problem of recovering a fully known matrix $\Y$ from its factorization, thus all entries are considered observed. 
We create three matrices in order to evaluate our approaches for recovery:
(a)~Random $10\times10$ matrix with $N=5$ that has $\GRR\leq 2$ (create by randomly generating $\taus$, $\U$, and $\V$),
(b)~Binary upper triangle matrix with size 10 ($\GRR$ of 1), and 
(c)~Band-diagonal matrix of size 10 and bandwidth 3, which has the linear rank of $8$ and $\GRR$ of 2.
Figure~\ref{fig:fullrec} presents the RMSE comparison of these three matrices as training progresses. For the upper triangle and the band diagonal, we fix threshold to $\tau=0.5$. The results show that Round works far better than others by converging to zero.  Moreover, linear approach is outperformed by the Multi-sigmoid without fixed thresholds in all, demonstrating it cannot recover even simple matrices. 

\paragraph{Matrix Completion}
Instead of fully-observed matrices, we now evaluate completion of the matrix when only a few of the entries are observed. 
We consider $50\times50$ upper-triangle and band-diagonal (bandwidth $10$) matrices, and sample entries from them, to illustrate how well our approaches can complete them. 
Results on held-out 20\% entries are given in Tables~\ref{tab:MC-UT} and \ref{tab:MC-BD}. 
In addition, we build a random matrix with size 50 and $\GRR$ 2, and present the results for this matrix in Table ~\ref{tab:MC-Ra}. 
As we can see, linear factorization in all three cases is outperformed by our proposed approaches. 
In band-diagonal, because of over-fitting of the Round approach, Multi-sigmoid performs a little better, and for upper-triangle, we achieve the best result for Round method by fixing $\tau=0.5$. 

\begin{figure*}[tb]
	\centering
	\begin{subfigure}{0.30\textwidth}
		\centering
		\includegraphics[width=\textwidth]{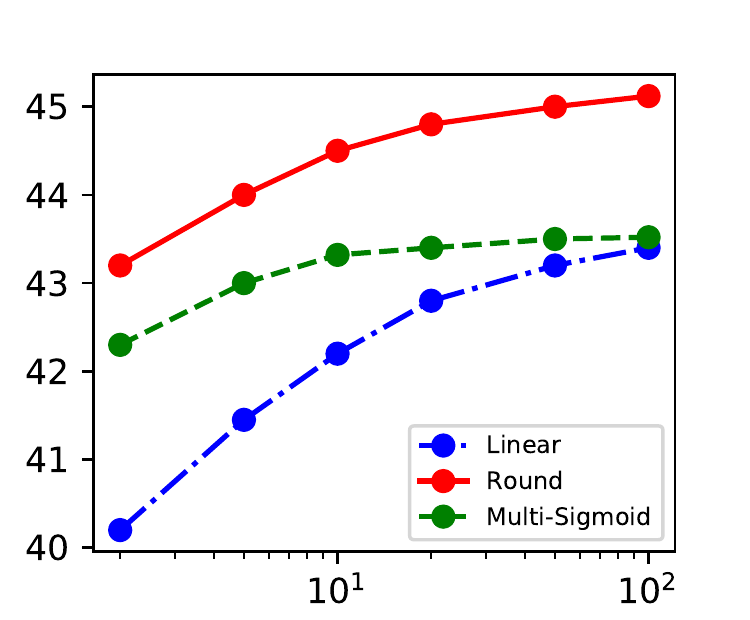}
		\caption{Percentage, smallnetfix}
	\end{subfigure}
	\begin{subfigure}{0.32\textwidth}
		\centering
		\includegraphics[width=\textwidth]{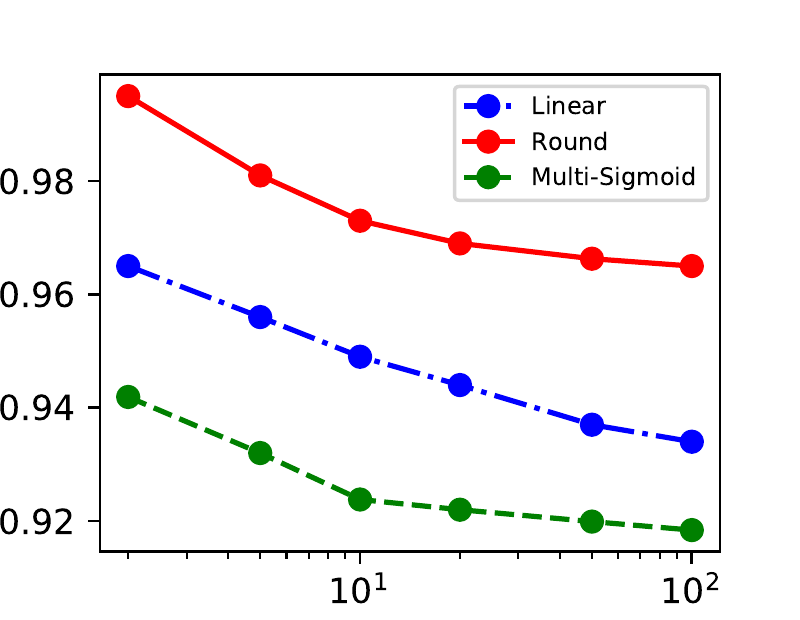}
		\caption{RMSE, smallnetflix}
	\end{subfigure}
	\begin{subfigure}{0.32\textwidth}
		\centering
		\includegraphics[width=\textwidth]{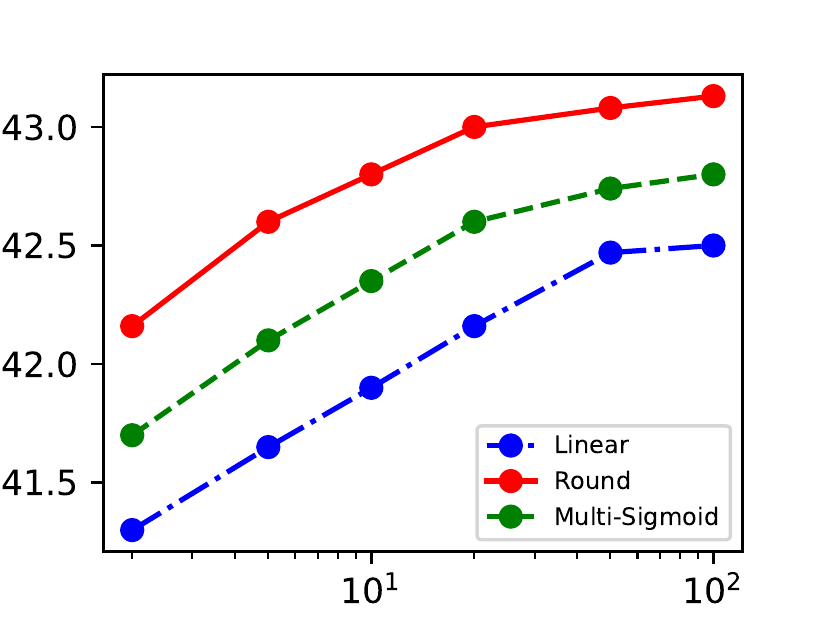}
		\caption{Percentage, movielens}
	\end{subfigure}
	
	\caption{Performance on recommendation datasets, as $k$ in increased}
	\label{fig:smallnet}
\end{figure*}

\paragraph{Matrix Completion on Real Data}
In this section we use the \emph{smallnetflix} movie ratings data for $95526$ users and $3561$ movies, where the training dataset contains $3,298,163$ ratings and validation contains $545,177$ ratings, while each one of ratings is an integer in $\{1,2,3,4,5\}$. 
We also evaluate on a second movie recommendation dataset, \emph{Movielens 100k}, with $100,000$ ratings from $1000$ users on $1700$ movies, with the same range as \emph{smallnetflix}.  
For this recommendation systems, in addition to RMSE, we also consider the notion of accuracy that is more appropriate for the task, calculated as the fraction of predicted ratings that are within $\pm0.5$ of the real ratings. As shown in Figure~\ref{fig:smallnet}, for \emph{smallnetflix}, linear factorization is better than Round approach from RMSE perspective, probably because linear is more robust to noise. On the other hand, Multi-sigmoid achieves better RMSE than linear method. Furthermore, both Round and Multi-sigmoid outperform the linear factorization in accuracy. \emph{Movielens} results for the percentage metric shows similar behavior as \emph{smallnetflix}, demonstrating that $\GRR$-based factorization can provide benefits to real-world applications. Furthermore, a comparison of our models with existing approaches on \emph{Movielens} dataset is provided in Table~\ref{tab:Ml}. We choose the RMSE result for smallest $k$ presented in those works. As we can see, our Multi-sigmoid method appear very competitive in comparison with other methods, while our Round approach result suffer from existence of noise in the dataset as before.   

\begin{table*}[tb]
	\centering
	\caption{Matrix completion for \textbf{Upper Triangular Matrices} ($k=1$)}
	\label{tab:MC-UT}
\begin{tabular}{lcccccccc}
\toprule
Proportion of Observations& 10\%& 20\%& 30\%& 40\%& 50\%& 60\%& 70\%& 80\% \\ 
\midrule
Linear&0.50&0.50&0.50&0.50&0.50&0.50&0.50&0.50\\ 
\addlinespace
Multi-Sigmoid&0.51&0.30&0.25&0.25&0.26&0.25&0.23&0.23\\ 
Multi-Sigmoid, $\tau=0.5$&0.58&0.37&0.36&0.36&0.35&0.35&0.34&0.34\\ 
Round&0.46&0.34&0.27&0.25&0.26&0.21&0.20&0.16\\ 
Round, $\tau=0.5$&\bf{0.38}&\bf{0.26}&\bf{0.23}&\bf{0.19}&\bf{0.15}&\bf{0.13}&\bf{0.15}&\bf{0.13}\\ 
\bottomrule
\end{tabular}
\end{table*}
\begin{table*}[tb]
	\centering
	\caption{Matrix completion for \textbf{Band Diagonal Matrices} ($k=2$)}
	\label{tab:MC-BD}
\begin{tabular}{lcccccccc}
\toprule
Proportion of Observations& 10\%& 20\%& 30\%& 40\%& 50\%& 60\%& 70\%& 80\% \\ \midrule
Linear&0.49&0.46&0.46&0.46&0.46&0.46&0.46&0.46\\ 
\addlinespace
Multi-Sigmoid&\bf{0.39}&\bf{0.26}&\bf{0.23}&\bf{0.23}&\bf{0.22}&\bf{0.21}&\bf{0.20}&\bf{0.20}\\ 
Multi-Sigmoid, $\tau=0.5$&0.48&0.49&0.33&0.31&0.30&0.29&0.29&0.29\\ 
Round&0.71&0.41&0.35&0.29&0.29&0.27&0.23&0.22\\ 
Round, $\tau=0.5$&0.61&0.57&0.39&0.52&0.58&0.30&0.29&0.34\\ \bottomrule
\end{tabular}
\end{table*}
\begin{table*}[tb]
	\centering
	\caption{Matrix completion with different number of samples for \textbf{Random low-GRR Matrices}}
	\label{tab:MC-Ra}
\begin{tabular}{lcccccccc}
\toprule
Proportion of Observations& 10\%& 20\%& 30\%& 40\%& 50\%& 60\%& 70\%& 80\% \\ \midrule
Linear&1.73&1.06&0.97&0.90&0.85&0.85&0.87&0.83\\ 
\addlinespace
Multi-Sigmoid&1.92&\bf{0.53}&\bf{0.48}&\bf{0.42}&\bf{0.39}&\bf{0.38}&0.36&0.35\\ 
Multi-Sigmoid (Fixed $\tau$)&1.96&1.54&1.37&1.32&1.29&1.28&1.25&1.23\\ 
Round&\bf{1.49}&0.92&0.60&0.48&0.48&0.39&\bf{0.30}&\bf{0.28}\\ 
Round (Fixed $\tau$)&2.44&1.50&1.50&1.43&1.36&1.39&1.44&1.34\\ \bottomrule
\end{tabular}
\end{table*}

\begin{table*}[tb]
	\centering
	\caption{RMSE on Movielens-100k for a variety of models with different low-rank approximation (k).}
	\label{tab:Ml}
\begin{tabular}{lcc}
\toprule
Models&Low-rank approximation&RMSE\\ \hline
APG~\citep{kwok2015accelerated}& k=70 & 1.037\\
AIS-Impute~\citep{kwok2015accelerated}& k=70 & 1.037\\
CWOCFI~\citep{lu2013second}& k=10 & 1.01\\
our Round& k=10 & 1.007\\
our Linear& k=10 & 0.995\\
UCMF~\citep{zhang2014information}& - & 0.948\\
our Multi-sigmoid& k=10 & 0.928\\
SVDPlusPlus~\citep{gantner2011mymedialite}& k=10 & 0.911\\
SIAFactorModel~\citep{gantner2011mymedialite}& k=10 & 0.908\\
GG~\citep{lakshminarayanan2011robust}& k=30 & 0.907\\
\bottomrule
\end{tabular}
\end{table*}








\section{Related Work}
\label{sec:related}

There is a rich literature on matrix factorization and its applications. 
To date, a number of link functions have been used, along with different losses for each, however here we are first to focus on expressive capabilities of these link functions, in particular of the ordinal-valued matrices~\citep{singh08:a-unified,koren2011ordrec,paquet2012hierarchical,udell14:generalized}.
\citet{nickel13:logistic} addressed tensor factorization problem and showed improved performance when using a sigmoid link function. 
\citet{marevcek2017matrix} introduced the concept of matrix factorization based on interval uncertainty, which results in a similar objective as our algorithm. 
However, not only is our proposed algorithm going beyond by updating the thresholds and supporting sigmoid-based smoothing, but we present results on the representation capabilities of the round-link function. 

A number of methods have approached matrix factorization from a probabilistic view, primarily describing solutions when faced with different forms of noise, resulting, interestingly, in link functions as well.
\citet{collins01:a-generalization} introduced a generalization of PCA method to loss function for non real-valued data,
such as binary-valued.
\citet{salakhutdinov08:bayesian} focused on Bayesian treatment of probabilistic matrix factorization, identifying the appropriate priors to encode various \emph{link} functions. 
On the other hand, \citet{lawrence09:non-linear} have analyzed non-linear matrix factorization based on Gaussian process and used SGD to optimize their model. 
However, these approaches do not explicitly investigate the representation capabilities, in particular, the significant difference in \emph{rank} when link functions are taken into account.

Sign-rank and its properties have been studied by \citet{nickel14:reducing,bouchard15:on-approximate,davenport14:1-bit}, and more recently, \citet{neumann2015some} provides in-depth analysis of round-rank. 
Although these have some similarity to $\GRR$, sign-rank and round-rank are limited to binary matrices, while $\GRR$ is more suitable for most practical applications, and further, we present extension of their results in this paper that apply to round-rank as well. 
Since we can view matrix factorization as a simple neural-network, research in understanding the complexity of neural networks~\citep{huang03:learning}, in particular with rectifier units~\citep{pan2016expressiveness}, is relevant, however the results differ significantly in the aspects of representation we focus on.
\section{Conclusions and Future Work}
\label{sec:conclusions}

In this paper, we demonstrated the expressive power of using link functions for matrix factorization, specifically the generalized round-rank ($\GRR$) for ordinal-value matrices.
We show that not only are there full-rank matrices that are low $\GRR$, but further, that these matrices cannot even be approximated by low linear factorization. 
Furthermore, we provide uniqueness conditions of this formulation, and provide gradient descent-based algorithms to perform such a factorization. 
We present evaluation on synthetic and real-world datasets that demonstrate that $\GRR$-based factorization works significantly better than linear factorization: converging faster while requiring fewer observations. 
In future work, 
we will investigate theoretical properties of our optimization algorithm, in particular explore convex relaxations to obtain convergence and analyze sample complexity.
We are interested in the connection of link-rank with different probabilistic interpretations, in particular, robustness to noise. 
Finally, we are also interested in practical applications of these ideas to different link functions and domains.


\bibliography{GRR}
\bibliographystyle{plainnat}


\newpage

\section*{Appendices}

\setcounter{section}{2}

\begin{thm:lemma}
For matrices $A ,B \in \{0,...,N\}^{n \times m}$:
\begin{align}
\GRR_{\tau_1,...\tau_N}(A) &\leq min(n,m)\\
 \GRR_{\tau_1,...\tau_N}(A) &=\GRR_{\tau_1,...\tau_N}(A^T)\\
 \GRR_{\tau_1,...\tau_N}(A+B) &\leq \GRR_{\tau_1,...\tau_N}(A)+\GRR_{\tau_1,...\tau_N}(B)
\end{align}
Where $+$ is in the real numbers and $A+B \in \{0,...,N\}^{n \times m}$.
\begin{proof}
According to definition of $\GRR$ and the fact that if $A=\GRF(C)$ then $ r(C) \leq min(n,m)$ we can conclude the first inequality. Furthermore, Since we know for any matrix $C$, $r(C)=r(C^T)$ and use the fact that if $A=\GRF(C)$ then $A^T=\GRF(C^t)$ we can show the second inequality as well. And the third inequality is the direct result of following famous inequality:
\begin{align}
r(A+b) &\leq r(A)+r(B)
\end{align}
\end{proof}
\end{thm:lemma}

\begin{thm:lemma}

the following decomposition holds for Generalized $\Round$ function: 
\begin{align}
\GRF_{\tau_1,...,\tau_N}(A)=\sum\limits_{i=1}^{N}\Round_{\tau_i}(A) 
\end{align}
\begin{proof}
Base on definition of Round Function $\sum\limits_{i=1}^{N}\Round_{\tau_i}(A)$ , counts the number of thresholds which are smaller than $A$, and this number is clearly equal to  $\GRF_{\tau_1,...,\tau_N}(A)$.
\end{proof}
\end{thm:lemma}

\begin{thm:lemma}
For any arbitrary subset of thresholds $T=\{\tau_{i_1},...,\tau_{i_r}\}$:
\begin{align}
\GRR_{\tau_1,...\tau_N}(A)\geq \GRR_{T}(\bar A)
\end{align}
Where $\bar A$ attained by the following transformation in matrix $A$:
\begin{align}
\bar A & =[b_{ij}]_{n\times m}\\
b_{ij} & =
\begin{cases}
	0, & \text{if } a_{ij} \in \{0,..,i_{1}-1\} \\
	1, & \text{if } a_{ij} \in\{i_{1},..,i_{2}-1\} \\
	\vdots\\
	r-1, & \text{if } a_{ij} \in\{i_{r},..,N-1\} 
\end{cases}
\end{align}
\begin{proof}
We define $\mathcal{B}$ and $\bar {\mathcal{B}}$ as follows:
\begin{align}
\mathcal{B} &=\{B|\GRF_{\tau_1,...,\tau_N}(B)=A\}\\
\bar {\mathcal{B}} &=\{\bar B|\GRF_{T}(\bar B)=A\} 
\end{align}
In result for any $B\in \mathcal{B}$, it is clear that $B \in \bar {\mathcal{B}}$
\end{proof}
\end{thm:lemma}

\begin{thm:lemma}

Following inequality holds for \GRR: 
\begin{align}
\GRR_{\tau_{1},...,\tau_{N}}(A)\leq \GRR_{\tau_1,...,\tau_N,\tau_{N+1}}(A) 
\end{align}
\begin{proof}
Similar to previous Lemma, if we define $\mathcal{B}$ and $\bar {\mathcal{B}}$ as follows:
\begin{align}
\mathcal{B} &=\{B|\GRF_{\tau_1,...,\tau_N}(B)=A\}\\
\bar {\mathcal{B}} &=\{\bar B|\GRF_{\tau_1,..,.\tau_N,\tau_{N+1}}(\bar B)=A\} 
\end{align}
Then it is clear that for any $\bar B\in \bar {\mathcal{B}}$, we have $\bar B \in \mathcal{B}$
\end{proof}
\end{thm:lemma}

\begin{thm:lemma}

Lets define the function $F: R^N \rightarrow N$ as follows:
\begin{align}
F(\tau_{1},...,\tau_{N})= \GRR_{\tau_1,...,\tau_N}(A) 
\end{align}
Where $A$ is a fix matrix in $\{0,...,N\}^{n\times m}$. Then we have the following inequality:
\begin{align}
F((\tau_{1}+\tau'_{1})/2,...,\tau_{N}) \leq F(\tau_{1},...,\tau_{N})+F(\tau'_{1},...,\tau_{N})
\end{align}
\begin{proof}
We define $\mathcal{B}$, $\mathcal{B}'$ and $\bar {\mathcal{B}}$ as follows:
\begin{align}
\mathcal{B} &=\{B|\GRF_{\tau_1,...,\tau_N}(B)=A\}\\
\mathcal{B}' &=\{B'|\GRF_{\tau_1',...,\tau_N}(B')=A\}\\
\bar {\mathcal{B}} &=\{\bar B|\GRF_{(\tau_{1}+\tau_{1}')/2,...,\tau_N}(\bar B)=A\} 
\end{align}
Accordingly, for any $B\in \mathcal{B}$ and $B' \in \mathcal{B}'$ we know $\frac{B+B'}{2}\in \bar {\mathcal{B}}$. Furthermore, since $\rank(\frac{B+B'}{2})=\rank(B+B')$ and $\rank(B+B') \leq \rank(B)+\rank(B')$ we can clearly prove the inequality.
\end{proof}
\end{thm:lemma}

\begin{thm:lemma}
We have the following inequality: 
\begin{align}
F(\tau_{1}+\tau'_{1},...,\tau_{N}+\tau'_{N}) \leq  F(\tau_{1},...,\tau_{N})+F(\tau'_{1},...,\tau'_{N})
\end{align}
\begin{proof}
Similar to previous Lemma, if we define $\mathcal{B}$, $\mathcal{B}'$ and $\bar {\mathcal{B}}$ as follows:
\begin{align}
\mathcal{B} &=\{B|\GRF_{\tau_1,...,\tau_N}(B)=A\}\\
\mathcal{B}' &=\{B'|\GRF_{\tau_1',...,\tau_N'}(B')=A\}\\
\bar {\mathcal{B}} &=\{\bar B|\GRF_{\tau_{1}+\tau_{1}',...,\tau_N+\tau_{N}'}(\bar B)=A\} 
\end{align}
For any $B\in \mathcal{B}$ and $B' \in \mathcal{B}'$ we know $B+B'\in \bar {\mathcal{B}}$. And since $\rank(B+B') \leq \rank(B)+\rank(B')$ we can clearly prove the inequality.
\end{proof}
\end{thm:lemma}

\begin{thm:thm}
For a given matrix $\Y \in \{0,\ldots,N\}^{n \times m}$, let's assume $\taus^*$ is the set of optimal thresholds, i.e. $ \GRR_{\taus^{\star}}(Y)=\text{argmin}_{\taus}\GRR_{\taus}(Y)$, then for any other $\taus'$: 
\begin{align}
\label{thresh_th_ap}
\GRR_{\taus'}(\Y) \leq N\times\GRR_{\taus^{\star}}(\Y)+1
\end{align}
\begin{proof}
To prove above inequality we first need two following lemmas:
\begin{thm:lemma}
\label{thresh_ap}
We have the following inequality for \GRR: 
\begin{align}
\GRR_{\tau_{1}+c,...,\tau_{N}+c}(\Y)\leq \GRR_{\tau_1,...,\tau_N}(\Y)+1 
\end{align}
Where $c$ is a real number.
\begin{proof}
We define $\mathcal{B}$ and $\mathcal{B}'$ as follows:
\begin{align}
\mathcal{B} &=\{B|\GRF_{\tau_1,..,.\tau_N}(B)=\Y\}\\
\mathcal{B}' &=\{B'|\GRF_{\tau_1+c,...,\tau_N+c}(B')=\Y\}
\end{align}
For an arbitrary $B\in \mathcal{B}$ let's assume we have matrix $\U$ and $\V$ in a way that, $B=\U \times \V^T$. If we add a column to the end of $\U$ and a row to the and of $\V$ and call them $\U'$ and $\V'$ as follows:
\begin{align}
U'&=\begin{bmatrix}
 & c\\ 
U & \vdots \\ 
 & c
\end{bmatrix},
&V'&=\begin{bmatrix}
 & 1\\ 
V & \vdots \\ 
 & 1 
\end{bmatrix}
\end{align}
It is clear that $B'=\U'\times \V'^T \in \mathcal{B}'$. Furthermore, by using the fact that $\rank(B')\leq \rank(B)+1$ we can complete the proof.
\end{proof}
\end{thm:lemma}

\begin{thm:lemma}
For arbitrary $k \in \R$, the following equality holds: 
\begin{align}
\GRR_{k\tau_{1},...,k\tau_{N}}(\Y) = \GRR_{\tau_{1},...\tau_{N}}(\Y)
\end{align}
\begin{proof}
Similar to previous Lemma, if we define $\mathcal{B}$ and $\mathcal{B}'$ as follows:
\begin{align}
\mathcal{B} &=\{B|\GRF_{\tau_1,...\tau_N}(B)=A\}\\
\mathcal{B}' &=\{B'|\GRF_{k\tau_1,...,k\tau_N}(B')=A\}
\end{align}
For any $B\in \mathcal{B}$  it is clear that  $k\times B\in \mathcal{B}'$. On the other hand, for any $B'\in \mathcal{B}'$  we know that  $ B'/k\in \mathcal{B}$.In result, by considering the fact that $\rank(kB) = \rank(B)$, we can complete the proof .
\end{proof}
\end{thm:lemma}

 base on These lemmas and the fact that for any $i \in \{1,...,N-1\}$, there exist an $\epsilon_i$ which will satisfies the following equality:
\begin{align}
\label{equal}
\GRR_{\tau_{1},...,\tau_{i}-\epsilon_i,...,\tau_{N}}(\Y) = \GRR_{\tau_{1},...\tau_{N}}(\Y)
\end{align}
We can show that there exists a set of  $\epsilon_i$ $(i \in \{1,...,N-1\})$, that transform $(\tau_{1},...\tau_{N})$ in to $(\tau'_{1},...,\tau'_{N})$ with a set of linear combinations. In another word, it means we have $k_0,...,k_{N-1}$ in a way that:
\begin{align}
T'=k_0T_0+...+k_{N-1}T_{N-1}
\end{align}
Where $T'=(\tau_{1}',...\tau_{N}')$, $T_0=(\tau_{1},...\tau_{N})$ and $T_i=(\tau_{1},...,\tau_{i}-\epsilon_i,...,\tau_{N})$ in vector format. Therefore, if we define ${\mathcal{B}}_i$ as follows:
\begin{align}
{\mathcal{B}}_i=\{B_i|\GRF_{T_i}(B_i)=A\}
\end{align}
And considering the fact that:
\begin{align}
 \rank(k_0B+...+k_{N-1}B_{N-1})&\leq \sum_{j=0}^{N-1} \rank(k_{j}B_{j})\\
  &=\sum_{j=0}^{N-1} \rank(B_{j}) 
\end{align}
Finally, with Lemma~\ref{thresh_ap} equation~\ref{equal} we can complete the theorem.
\end{proof}
\end{thm:thm}

\stepcounter{section}
\setcounter{section}{3}
\begin{thm:lemma}
For any matrix $A$, if there exists a submatrix $A'$ in a way that $\rank(A')=R$ and $\GRR_{\taus}(A')=r$, then $\GRR_{\taus}(A)\geq  r$ and $\rank(A)\geq R$
. \begin{proof}
If we consider the linear rank as the number of independent row (column) of the matrix, consequently having a rank of $r$ for submatrix $A'$ means there exist at least $r$ independent rows in matrix a $A$. Using this argument we can simply prove above inequalities.
\end{proof}
\end{thm:lemma}

\begin{thm:thm}
		If a matrix $\Y\in\R^{n\times m}$ contains $k$ rows, $k\leq n,k\leq m$, such that $R=\{Y_{R_1},...,Y_{R_k}\}$, two columns $C=\{j_0,j_1\}$, and:
	\begin{enumerate}
	\item rows in $R$ are distinct from each other, i.e, $\forall i,i'\in R, \exists j,Y_{ij}\neq Y_{i'j}$,
	\item columns in $C$ are distinct from each other, i.e, $\exists i,Y_{ij_0}\neq Y_{ij_1}$, and
	\item matrix spanning $R$ and $C$ are non-zero constants, w.l.o.g. $\forall i\in R,Y_{ij_0}=Y_{ij_1}=1$,
	\end{enumerate}
	then $\rank(\Y)\geq k$.
	\begin{proof}
	
				Let us assume $\rank(\Y)<k$, i.e. $\exists k'<k, \U\in\R^{k'\times n}, \V\in\R^{k'\times m}$ such that $\Y=\U^T\times\V$.
		Since the rows $R$ and the columns in $C$ are distinct, their factorizations in $\U$ and $\V$ have to also be distinct, i.e. $\forall i,i'\in R, i\neq i', \U_{i}\neq\U_{i'}$ and $\V_{j_0}\neq\V_{j_1}$.
		Furthermore, $\forall i,i'\in R, i\neq i', \not\exists a,\U_{i}=a\U_{i'}$ and $\not\exists a,\V_{j_0}=a\V_{j_1}$ for $a\neq0$, it is clear that $\U_i\cdot\V_{j_0}=\U_i\cdot\V_{j_1}=1$ (and similarly for $i,i'\in R$).

		Now consider a row $i\in R$. Since $\forall j\in C, \Y_{ij}=1$, then $\U_i\cdot\V_{j}=1$.
		As a result, $\V_j$ are distinct vectors that lie in the hyperplane spanned by $\U_i\cdot\V_j=1$.
		In other words, the hyperplane $\U_i\cdot\V_j=1$ defines a $k'$-dimensional hyperplane tangent to the unit hyper-sphere.

		Going over all the rows in $R$, we obtain constraints that $\V_j$ are distinct vectors that lie in the intersection of the hyperplanes spanned by $\U_i\cdot\V_j=1$ for all $i\in R$.
		Since all $\U_i$s are distinct, there are $k$ distinct $k'$-dimensional hyperplanes, all tangent to the unit sphere, that intersect at more than one point (since $\V_j$s are distinct).

		Since $k$ hyper-planes tangent to unit sphere can intersect at at most one point in $k'<k$ dimensional space, $\V_j$ cannot be distinct vectors. Hence, our original assumption $k'<k$ is wrong, therefore, $\rank(\Y)\geq k$.
	\end{proof}
\end{thm:thm}

\stepcounter{section}
\setcounter{section}{4}
\begin{thm:thm} (Necessary Condition)
For a target matrix $\Y \in \mathbb{V}^{n \times m}$ with few observed entries and given $\GRR(\Y)=r$, we consider set of $\{\Y_{i_1j},...,\Y_{i_rj}\}$ to be the $r$ observed entries in an arbitrary column $j$ of $Y$. 
Given any matrix $\X \in \mathcal{X}^{\star}$, $\X=\U\times \V^{T}$, and taking an unobserved entry $\Y_{ij}$, we define $a_{i_kj}$ as: 
$
\U_i=\sum_{k=1}^{r} a_{i_kj} \U_{i_k} 
$, where $\U_d$ ($d \in \{1,...,n\}$) is the $d^\text{th}$ row of matrix $\U$ and $i_k$ represents the index of observed entries in $j$th column.
Then, the necessary condition of uniqueness of $\Y$ is:
\begin{align}
\sum_{k=1}^{r}\left | a_{i_kj} \right | \leq  \epsilon\left(\frac{T_\text{min}}{T_\text{max}}\right)
\end{align}
Where $r=\GRR(\Y)$, $T_\text{min}$ and $T_\text{max}$ are the length of smallest and largest intervals and $\epsilon$ is a small constant. 
\begin{proof}
To better understand the concept of uniqueness in $\GRR$ benchmark, let's first look at the uniqueness in fixed value matrix factorization (traditional definition(MF)).

 In fixed value matrix factorization, it is proved that to achieve uniqueness, we need at least $r=\rank(\X)$ observation in each column(other than the independent columns). Therefore, if we decompose $\X$ as $\X=\U\V^T$, and plan to changed only unobserved entries of $\Y$ in column $j$ (in opposed to uniqueness), we need to change the $jth$ row of matrix $\V$. To do so, let's assume we change the $jth$ row to:
 \begin{align}
[\V_{j1}+c_1,...,\V_{jr}+c_r]
\end{align}
Now since we know $\rank(U)=r$ and assume the respective rows of $\U$ to observed entries of column $j$ in matrix $\X$ are independent (this is a required assumption for uniqueness), we can show that only possible value for $c_1,..., c_r$ which does not change the observed entries of $\X$ is $0$, which confirm the uniqueness.  

The biggest difference between MF based on $\GRR$ and traditional MF is the fact that the observed entries of matrix $\X$ are not fixed in $\GRR$ version, and can change through the respective interval. In result, to achieve uniqueness we need to find a condition which for any column of $\X$, by changing respective row of $\V$, while the value of observed entries stay in the respected intervals, the value of unobserved ones wouldn't change dramatically which result in moving to other intervals. To do so, we will calculate the maximum of the possible change for an arbitrary unobserved entry of column $j$ in matrix $\Y$.

Let's call the $r$ observed entries of column's $j$ of matrix $\Y$, $\Y_{i_{1}j},...,\Y_{i_{r}j}$. Similar to MF case, we assume that the respective rows of $\U$ to these entries are independent. In result, if we represent the change in entries of $jth$ rows of $\V$ by $c_i$, we should have:
\begin{align}
\begin{bmatrix}
\U_{i_1}\\ 
\vdots \\ 
\U_{i_r}
\end{bmatrix}
\times
\begin{bmatrix}
c_1\\ 
\vdots \\ 
c_r
\end{bmatrix}
=
\begin{bmatrix}
\epsilon_{i_1j}\\ 
\vdots \\ 
\epsilon_{i_rj}
\end{bmatrix}
\end{align}            
Where $\U_{i_k}$ is the $i_k th$ row of $\U$, and $\epsilon_{i_kj}$ is the possible change for $\X_{i_{k}j}$, based on the observed interval. Therefore:
\begin{align}
\epsilon_{i_kj} \in (\tau_{i_kj}\downarrow-\X_{i_kj},\tau_{i_kj}\uparrow-\X_{i_kj})=(\epsilon_{i_kj}^-,\epsilon_{i_kj}^+)
\end{align}
Now let's assume we want to find the maximum possible change for $\X_{sj}$ considering that $\Y_{sj}$ is and unobserved entry. Since $\U_{i_k}$'s are independent, there exist $a_1,..a_r$ which:
\begin{align}
\U_s=\sum_{k=1}^{r}a_{i_kj}\U_{i_k}
\end{align}
Therefore, we can show the change in entry $\X_{sj}$ as:
\begin{align}
A=\sum_{k=1}^{r}a_{i_kj}\epsilon_{i_kj}
\end{align}
In result, for the maximum possible change we have:
\begin{align}
max|A|=max(\sum_{k=1}^{r}a_{i_kj}\epsilon_{i_kj}^{sign(a_{i_kj})},|\sum_{k=1}^{r}a_{i_kj}\epsilon_{i_kj}^{-sign(a_{i_kj})}|)
\end{align}
Where $sign(.)$ is the sign function. On the other hand we know:
\begin{align}
\sum_{k=1}^{r}a_{i_kj}\epsilon_{i_kj}^{sign(a_{i_kj})}+|\sum_{k=1}^{r}a_{i_kj}\epsilon_{i_kj}^{-sign(a_{i_kj})}|=\sum_{k=1}^{r}|a_{i_kj}|T_{i_kj}
\end{align}
\begin{align}
\Rightarrow max|A| \geqslant \frac{1}{2}\sum_{k=1}^{r}|a_{i_kj}|T_{i_kj}
\end{align}
 Where $T_{i_kj}$ is the length of the interval entry of $\mathbf{\bar X}_{i_kj}$. Clearly, to achieve the uniqueness we need $max|A|\leq T_{sj}$. But, since the entry $\X_{sj}$ is unobserved we don't know the value of $T_{sj}$. In result, for sake of uniqueness in the worst case we need:
\begin{align}
\sum_{k=1}^{r}|a_{i_kj}|T_{max} &\leq \epsilon T_{min}\\
\Rightarrow\sum_{k=1}^{r}|a_{i_kj}| &\leq\epsilon \frac{T_{min}}{T_{max}}
\end{align} 
Where $T_{min}$ and $T_{max}$ are the smallest and the biggest interval, and $\epsilon$ is a small real constant. 
 \end{proof}
\end{thm:thm}

\begin{thm:thm} (Sufficient Condition)
Using above necessary condition, for any unobserved entry $\Y_{ij}$ of matrix $\Y$ we define $\bar\epsilon$ as minimum distance of $\X_{ij}$ with its respected interval's boundaries. Than, we will have the following inequality as sufficient condition of uniqueness:
\begin{align}
\bar\epsilon \geq \max (\sum_{k=1}^{r}a_{i_kj}\epsilon_{i_k}^{\text{sign}(a_{i_kj})} , \left | \sum_{k=1}^{r}a_{i_kj}\epsilon_{i_k}^{-\text{sign}(a_{i_kj})}  \right |)
\end{align}
Where $r$ and $a_{i_kj}$ are defined as before, $\epsilon_{i_kj}^{+}$ is defined as the distance of $\X_{i_kj}$ with its respected upper bound and $\epsilon_{i_kj}^{-}$ is defined as negative of the distance of $\X_{i_kj}$ with its respected lower bound.
\begin{proof}
Sufficient condition is the direct result of Necessary Conditions proof. By combining (48) with the definition of uniqueness we can achieve the Sufficient Condition.
\end{proof}
\end{thm:thm}

\begin{thm:thm}
For any $\epsilon>0$ and matrix $\Y$, $\text{sign-rank}(\Y) = \apprank_\sigmoid(\Y)$.
\begin{proof}
Let $\Bm_\sigmoid^\epsilon(k)=\{\B\in\{0,1\}^{n\times m}; \apprank_\sigmoid(\B)=k\}$, i.e. the set of binary matrices whose $\apprank_\sigmoid$ is equal to $k$, and $\Bm_\step(k)=\{\B\in\{0,1\}^{n\times m}; \text{sign-rank}(\B)=k\}$.  We prove the theorem by showing both directions.
\textul{$\Bm_\step\subseteq\Bm_\sigmoid$}:
Any $\U,\V$ that works for $\step$ should work with $\sigmoid$ if multiplied by a very large number, i.e. take a sufficiently large $\eta$, and $\U_\sigmoid=\eta\U_\step,\V_\sigmoid=\eta\V_\step$.
Then, $\X_\sigmoid=\eta^2\X_\step$ and if we set $\lparams_\sigmoid=\eta^2\lparams_\step$, then $(\X_\sigmoid-\lparams_\sigmoid)=\eta^2(\X_\step-\lparams_\step)$, therefore will have the same sign, and $\Y_\sigmoid=\sigmoid(\X_\sigmoid)$ will be arbitrarily close to $0$ and $1$ in $\Y_\step$.
\textul{$\Bm_\sigmoid\subseteq\Bm_\step$}:
Any $\U,\V$ that works for $\sigmoid$ will directly work with $\step$.
\end{proof}
\end{thm:thm}

\begin{thm:rmk}
To extend Theorem 4.3 to multi-ordinal cases, we need to show that for any arbitrary set of thresholds in GRR, there exists another set of thresholds for multi-sigmoid function which will satisfy the condition in theorem 4.3 for multi-ordinal matrices. The procedure of proof is similar to binary cases. The only difference is the fact that after multiplying our matrices into a big enough constant, we need to choose multi-sigmoid’s thresholds in a way that will guarantee the multi-sigmoid(X) is close enough of to GRF(X)(which is equal to Y). 
\end{thm:rmk}

\end{document}